\documentclass[letterpaper]{article} 
\usepackage{aaai25}  
\usepackage{times}  
\usepackage{helvet}  
\usepackage{courier}  
\usepackage[hyphens]{url}  
\usepackage{graphicx} 
\usepackage{natbib}  
\usepackage{caption} 
\usepackage{algorithm}
\usepackage{algorithmic}
\usepackage{amsthm,bm}
\usepackage{amssymb}
\usepackage{amsfonts}
\usepackage{amsmath}
\usepackage{subcaption}
\usepackage{newfloat}
\usepackage{listings}
\DeclareCaptionStyle{ruled}{labelfont=normalfont,labelsep=colon,strut=off} 
\floatstyle{ruled}
\newfloat{listing}{tb}{lst}{}
\floatname{listing}{Listing}
\usepackage{caption}
\theoremstyle{plain}

\theoremstyle{definition}

\theoremstyle{remark}

\def \E {\mathop{{\mathbb{E}}}\limits}

\def \N {\mathcal{N}}

\def \D {\mathbb{D}}

\def \cb {\mathbf{c}}

\def \zb {\mathbf{z}}
\def \xb {\mathbf{x}}
\def \yb {\mathbf{y}}

\def \varepsilonb {\bm{\varepsilon}}
\def \thetab {\bm{\theta}}
\def \betab {\bm{\beta}}
\def \taub {\bm{\taub}}

\def \E {\mathop{{\mathbb{E}}}\limits}

\def \N {\mathcal{N}}

\def \D {\mathcal{D}}

\def \Em {\mathcal{E}}

\def \Ib {\mathbf{I}}

\title{Unlearning Concepts from Text-to-Video Diffusion Models}
\author{
    Shiqi Liu\textsuperscript{\rm 1},
    Yihua Tan\textsuperscript{\rm 1}\thanks{Corresponding author.},
}
\affiliations{
    \textsuperscript{\rm 1}Huazhong University of Science and Technology,Wuhan 430074,PR China\\

    shiqi.liu647@foxmail.com, yhtan@hust.edu.cn
}

\AddToHook{cmd/@maketitle/after}{\ADDINITIALFIGURE}
\newcommand{\ADDINITIALFIGURE}{%
      \includegraphics[width=\textwidth]{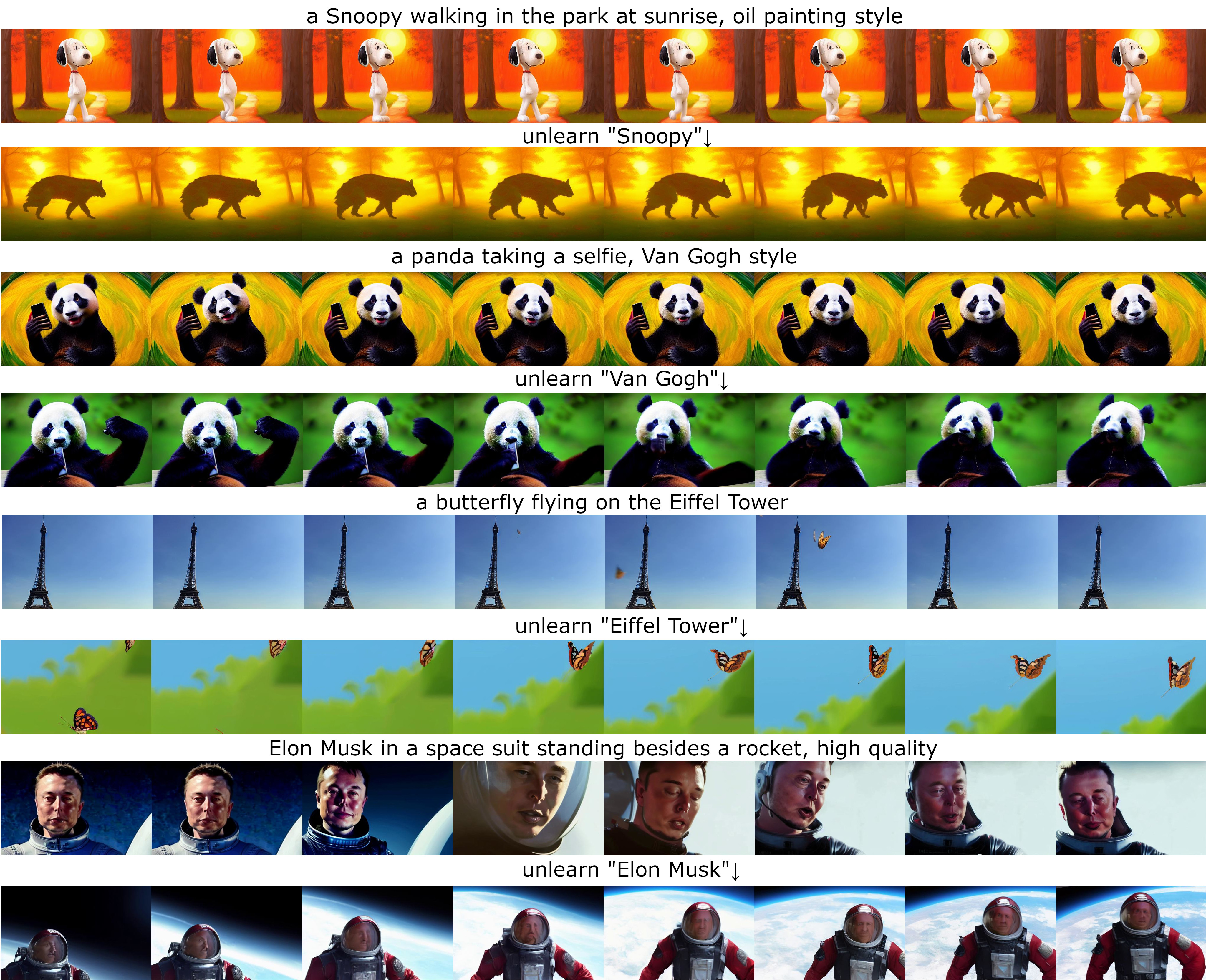}%
    \setcounter{figure}{0}
    \captionof{figure}{The comparison of the concept-preserved and concept-unlearned videos generated by our algorithm.\label{fig:unlearnmotage}}
  }
\usepackage{bibentry}
\begin{document}

\maketitle

\begin{abstract}
With the advancement of computer vision and natural language processing, text-to-video generation, enabled by text-to-video diffusion models, has become more prevalent.  These models are trained using a large amount of data from the internet. However, the training data often contain copyrighted content, including cartoon character icons and artist styles, private portraits, and unsafe videos.  Since filtering the data and retraining the model is challenging, methods for unlearning specific concepts from text-to-video diffusion models have been investigated. However, due to the high computational complexity and relative large optimization scale, there is little work on unlearning methods for text-to-video diffusion models. We propose a novel concept-unlearning method by transferring the unlearning capability of the text encoder of text-to-image diffusion models to text-to-video diffusion models. Specifically, the method optimizes the text encoder using few-shot unlearning, where several generated images are used.  We then use the optimized text encoder in text-to-video diffusion models to generate videos. Our method costs low computation resources and has small optimization scale. We discuss the generated videos after unlearning a concept.  The experiments demonstrates that our method can unlearn copyrighted cartoon characters, artist styles, objects and people's facial characteristics. Our method can unlearn a concept within about 100 seconds on an RTX 3070.  Since there was no concept unlearning method for text-to-video diffusion models before, we make concept unlearning feasible and more accessible in the text-to-video domain.
\end{abstract}

%

\section{Introduction}
Recent text-to-video diffusion generative models\cite{wang2023lavie,ho2022imagen,yin2023nuwa} have attracted attention because of their outstanding video quality, stable learning procedure, and seemingly infinite generation capabilities, surpassing the previous state-of-art generative adversarial networks\cite{goodfellow2020generative,goodfellow2014generative}. Classifier-free guidance\cite{ho2021classifier} allows us generate high-quality videos on the basis of natural language input. These models are able to imitate a wide range of concepts since they are trained on vast internet datasets. 

Their ability to imitate potentially copyrighted content is a major concern regarding text-to-video models. They can faithfully generate copyrighted videos such as ``Snoopy," an iconic beagle dog from the beloved comic strip Peanuts, as shown in Figure 1. The AI-generated art is on par with human-generated art. Another issue is that the models can faithfully replicate an artist's style. Users of large-scale text-to-video generation systems can use prompts including ``in the style of [artist]" to mimic the styles of specific artists, which may reduce the value of the original work. The Van Gogh-style video ``a panda taking a selfie" is shown in Figure 1. Some artists have sued the makers and providers of certain generation models, raising new legal issues \cite{setty2023ai}.

Apart from copyright infringement issues, privacy and safety are other major concerns. Text-to-video diffusion models can generate specific facial characteristics through text prompts that include names, if the training datasets contain corresponding videos of those names. An example of this is shown in Figure 1: a generated video of Elon Musk. However, this generation of facial characteristics may raise concerns about privacy and portrait rights as outlined in the Civil Code of the People's Republic of China. Additionally, malicious use of these generated videos could contribute to the spread of fake news and misinformation. Text-to-video diffusion models are also capable of generating nude and pornographic videos. These concerns all necessitate technical solutions.

\begin{figure}
  \centering
  \includegraphics[width=0.45\textwidth]{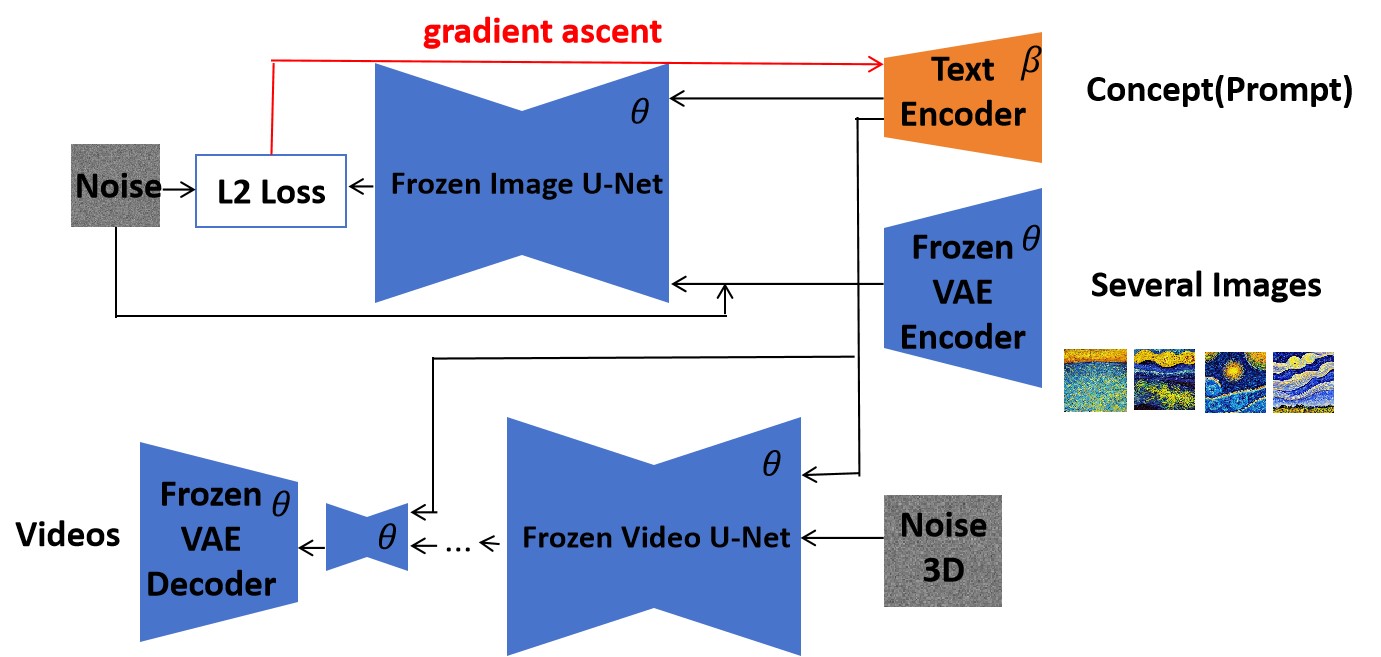}
  \caption{Overview of our proposed method. $\thetab$
denotes that parameters are fixed and $\betab$ denotes that parameters need to be optimized. }\label{fig:overview}
\end{figure}
Cleaning the datasets and retraining the text-to-video models require a great amount of work and expenditure. For example, training text-to-video diffusion models, as described in \cite{wang2023lavie,ho2022imagen}, requires about 10 million videos. One feasible approach is to use unlearning methods \cite{bourtoule2021machine}, which are proposed to eliminate the influence of specific data or concepts. Previous studies have successfully unlearned specific concepts from text-to-image models by optimizing the weights of U-Net \cite{ronneberger2015u}, the generative module \cite{gandikota2023erasing,kumari2023ablating,gandikota2024unified,zhang2024forget,zhao2024separable}. A problem with this method is that optimizing the parameters of U-Net can lead to a decline in generation quality. Another alternative is to unlearn specific concepts from text-to-image models by optimizing the parameters of the text encoder \cite{radford2021learning,raffel2020exploring}. This method \cite{fuchi2024erasing} uses gradient ascent of parameters with regard to the concept to be unlearned on the images related to the concepts. However, due to the high computational complexity and relative large optimization scale, there is little work on unlearning methods for text-to-video diffusion models.  Since some text-to-image diffusion models \cite{rombach2022high} and text-to-video diffusion models \cite{wang2023lavie} share the same text encoder, it is natural to question whether we can transfer the unlearning capability of text-to-image diffusion models to text-to-video diffusion models.

\begin{figure*}
\centering

\begin{subfigure}{0.45\textwidth}
        \centering
        \includegraphics[width=\linewidth]{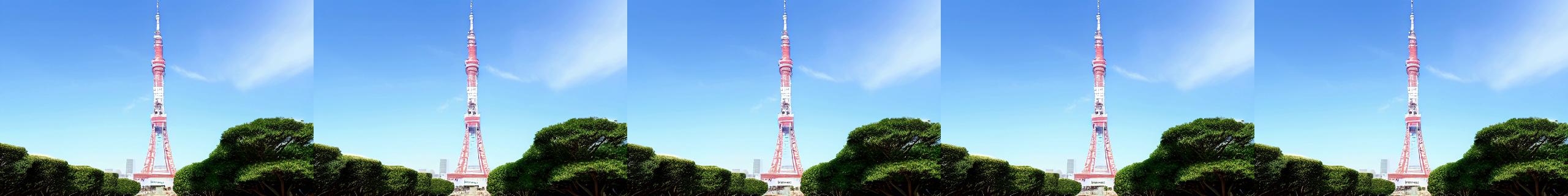}
        \caption{a cat climbing on Tokyo Tower}
        \label{fig:subfig1}
    \end{subfigure}
\begin{subfigure}{0.45\textwidth}
        \centering
        \includegraphics[width=\linewidth]{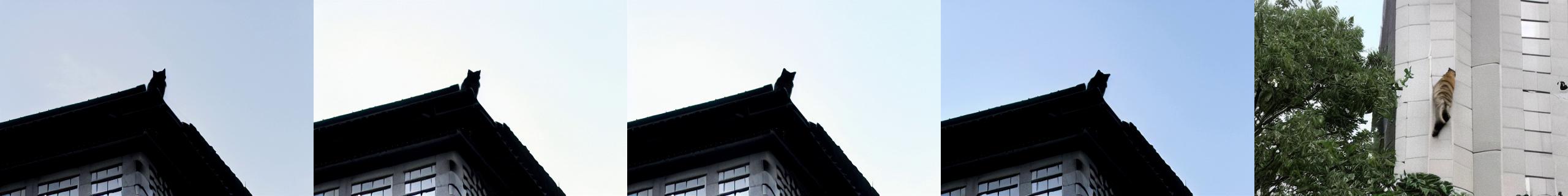}
        \caption{ Unlearn "Eiffel Tower"}
        \label{fig:subfig2}
    \end{subfigure}
\begin{subfigure}{0.45\textwidth}
        \centering
        \includegraphics[width=\linewidth]{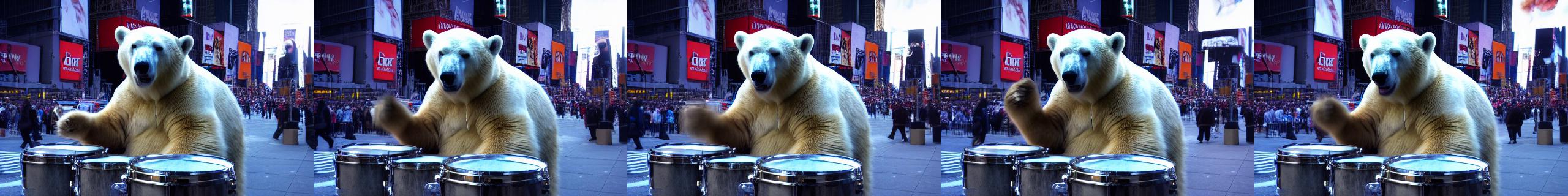}
        \caption{  a polar bear playing drum kit in NYC Times Square, 4k, high resolution}
        \label{fig:subfig3}
    \end{subfigure}
\begin{subfigure}{0.45\textwidth}
        \centering
        \includegraphics[width=\linewidth]{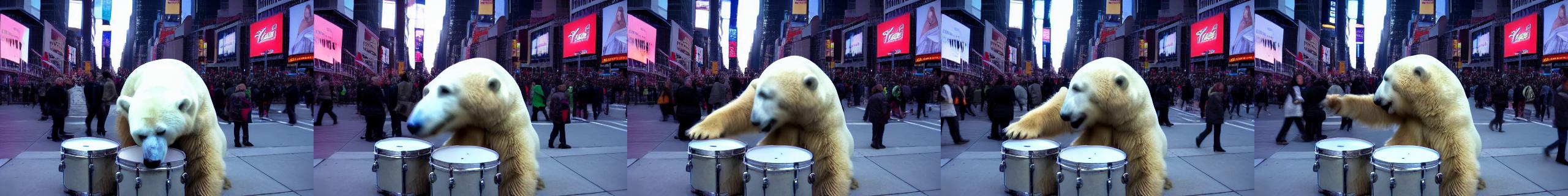}
        \caption{ Unlearn "Eiffel Tower"}
        \label{fig:subfig4}
    \end{subfigure}
\begin{subfigure}{0.45\textwidth}
        \centering
        \includegraphics[width=\linewidth]{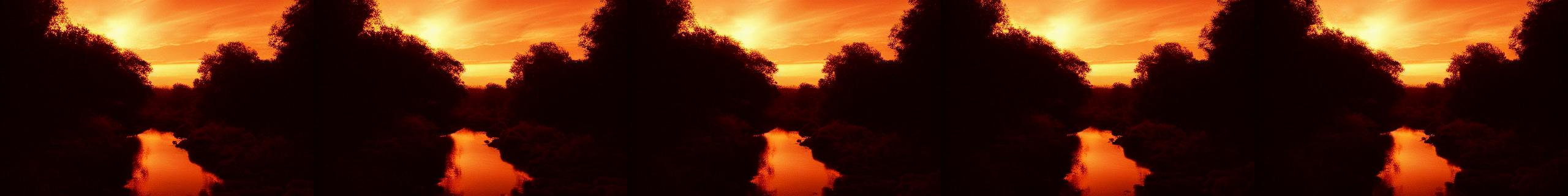}
        \caption{  jungle river at sunset, ultra quality}
        \label{fig:subfig3}
    \end{subfigure}
\begin{subfigure}{0.45\textwidth}
        \centering
        \includegraphics[width=\linewidth]{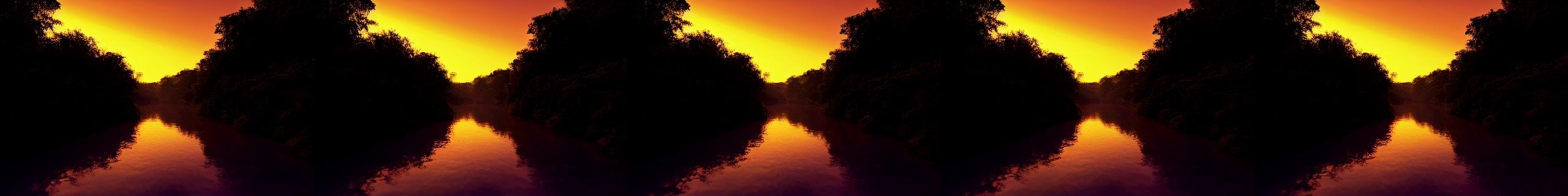}
        \caption{ Unlearn "Eiffel Tower"}
        \label{fig:subfig4}
    \end{subfigure}
\caption{The comparison of the effects on other concepts after unlearning "Eiffel Tower". \label{fig:other effects} }
\end{figure*}

We propose a method, based on text-to-image diffusion models \cite{fuchi2024erasing}, for unlearning specific concepts without altering the video-generating module. We aim to achieve unlearning by altering the embedding of text conditioning in the text encoder. The text-to-video domain unlearning is with high computational complexity and requires relative large optimization scale. We address the problem by using the same text encoder to transfer the unlearning capability of text-to-image diffusion models to text-to-video diffusion models. The transfer learning confines the computational complexity into the text-to-image level and reduces the optimization scale.  We utilize a few images of the concept to make small changes to remove the concept from the text encoder in the image domain and reuse the text encoder in the text-to-video diffusion models to remove the concept in the video domain. Since there are only minor parameter adjustments using image domain optimization on the text encoder, it operates quickly with small costs compared to unlearning in the text-to-video domain. According to Figure~1, it is evident that our method can successfully unlearn concepts.

Our contributions are the following:
\begin{itemize}
\item The unlearning procedure transfers from text-to-image diffusion models to text-to-video diffusion models.
\item The unlearning procedure takes only about 100 seconds on an RTX 3070.
\item Concept unlearning is achieved by providing a few images regarding the concept to be unlearned without using videos or optimizing U-Net.
\end{itemize}

\section{Related Works}
\subsection{Text-to-Video Diffusion Models and Text Encoder}
Due to the their simple training objective, the stable training processes, and good generation performance, denoising diffusion models\cite{ho2020denoising} have been successful in image generation\cite{rombach2022high} and have gradually attracted more attentions on video generation\cite{wang2023lavie}. Some representative works include Imagen video\cite{ho2022imagen}, which uses a cascade sampling pipeline for video generation,  NUWA-XL\cite{yin2023nuwa}, which uses a ``coarse-to-fine" generation pipeline, and Lavie\cite{wang2023lavie}, which uses a cascade latent diffusion generation pipeline.

Text-to-video generation utilizes a text encoder to encode the prompt, providing a semantic condition for video generation. There two commonly used text encoders. One is CLIP\cite{radford2021learning}. The other is T5\cite{raffel2020exploring,ni2021sentence}. CLIP is trained on 400 million (image,text) pairs collected from the internet\cite{radford2021learning}. The author compared its performance against over 30 different existing computer vision datasets and the model is competitive with supervised baseline\cite{radford2021learning}. T5 performs well on sentence transfer tasks and is state-of-art sentence embedding model\cite{raffel2020exploring}. In many works, the text encoder is shared and fixed across the video and image domain\cite{wang2023lavie}\cite{rombach2022high}.
 
 \cite{fuchi2024erasing}'s work showed that we can unlearn the concept by adjusting only the parameters of the text encoder in text-to-image diffusion models. Our method builds upon the work of \cite{fuchi2024erasing}. By utilizing the text-to-image and text-to-video models that shared the same text encoder, the unlearning effect on the text encoder on text-to-image domain can be transferred to the text-to-video domain. 
 
\subsection{Memorization and Unlearning}
While the original goal of machine learning is to generalize rather than memorize, large diffusion models are capable of both exact memorization(as shown in\cite{gu2023memorization}) and unintentional memorization\cite{gu2023memorization}. The memorization phenomenon could lead to copyright and privacy issues, prompting the development of unlearning techniques. Unlearning aim to eliminate the influence of specific training samples on the models by adjusting their parameters. This essentially makes the model behave as if it had never encountered those samples. Some unlearning approaches focus on unlearning specific training samples, while other target broader concepts \cite{ma2024dataset}, as in our work. Our work corresponds to the latter unlearning purpose, aiming to unlearn the high level concepts such as artist styles, concrete copyright icons, and individuals's appearances.

\subsection{Unlearning Concepts from Text-to-Image Diffusion and Text-to-Video Diffusion}
Many video-generation and image-generation models are trained on a massive amount of data on the Internet\cite{radford2021learning}. As a result, these datasets often contain copyrighted material and privacy-senstitive content, which causes problems when deploying the models to the market. Cleaning the data and retraining the model is one method to fix the issue. However, the cost of retraining a model trained on the datasets with millions samples is often unaffordable. The more practical method is to unlearn some data or concepts of the baseline models\cite{bourtoule2021machine}.

 \begin{figure*}
  \centering
  \includegraphics[width=\textwidth]{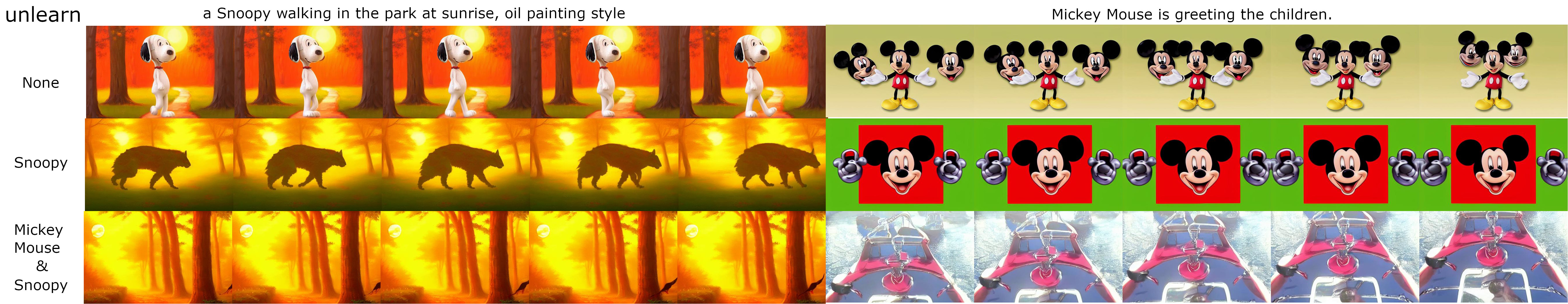}
  \caption{The comparison of the effects when unlearning multiple concepts}\label{fig:mutipleconceptunlearning}
\end{figure*}

There are several works on unlearning concepts from text-to-image diffusion models. Most of them\cite{gandikota2023erasing,kumari2023ablating,gandikota2024unified,zhang2024forget,zhao2024separable} focus on the generation module, the U-Net.
Specifically, \cite{gandikota2023erasing,kumari2023ablating,zhao2024separable} focus on updating the entire U-Net parameters, while \cite{gandikota2024unified,zhang2024forget} focus on updating the cross-attention part of the U-Net parameters. \cite{fuchi2024erasing} focus on updating the parameters of the text encoder. Updating the U-Net parameters comes with high computational complexity and is time-consuming. Additionally, updating the U-Net may influence the delicate generation and further affect the fidelity of the generated image. In contrast, updating the text encoder does not affect the fidelity and is time-efficient. Since are some text-to-image and text-to-video models utilize a shared text encoder\cite{wang2023lavie,rombach2022high}, there is great potential to transfer the unlearning effect from image domain to the video domain.

Currently, there is a scarcity of research on unlearning concepts from text-to-video diffusion models. In our method, we leverage the transfer capability of the shared text encoder within text-to-video models.
\subsection{Unlearning in Transformer-based Models}
Several methods implement the unlearning in transformed-based models. \cite{chen2023unlearn} introduced unlearn layers into the large language models to achieve unlearning. \cite{tian2024forget} utilize gradient information to address excessive unlearning in large language model. 

Compare to \cite{chen2023unlearn}'s work, Our method does not introduce auxiliary parameters into the unlearning model. The method proposed by \cite{tian2024forget} to overcome excessive unlearning is promising for the future work on our method.

\section{Methodology}

We aim to prevent the generation of specific concepts in video diffusion models. To achieve this, we propose transferring the unlearning capability of text-to-image diffusion models to text-to-video models. Specifically, we select video and image diffusion models that share a common text encoder. We then implement unlearning on the text encoder in the text-to-image domain to transfer the unlearned ability to the text-to-video domain. Importantly, we fix the U-Nets of both models, ensuring that the models' generation capability is preserved.

Our method builds upon \cite{fuchi2024erasing}'s work. In their work, the author only update the parameters of the text encoder, and they lists the reasons. First, research suggests that the quality of text-image alignment correlates with the quality of the text encoder\cite{saharia2022photorealistic}. DALLE-3\cite{betker2023improving} successfully achieves high-quality generation by choosing GPT4\cite{achiam2023gpt} as its text encoder. Second, the output of text encoder is a multi-dimensional vector that preserves meaningful information about the concepts described in the text.

In addition to the benefit discussed above, the transferability of the text encoder is also important. Text encoders trained on the text-to-image and text-to-video domains may play different roles in semantic embedding. Specifically, one role is for image semantic embedding, and the other is for video semantic embedding. However, we postulate that unlearning concepts in the text encoder can have transfer capability to both its text-to-image and text-to-video domain semantic embedding.

\begin{figure*}[h]
  \centering
  \begin{minipage}[t]{\textwidth}
  \centering
    \includegraphics[width=\textwidth]{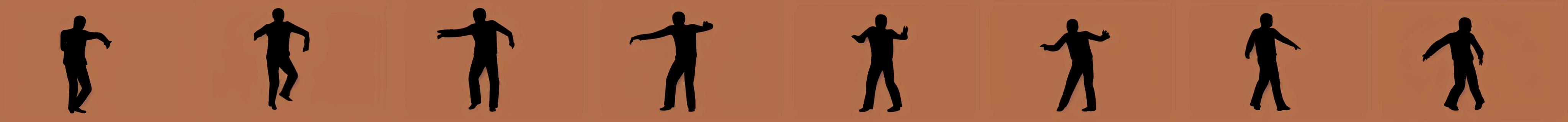}
    {\small A person is dancing jazz.}
  \end{minipage}
  \begin{minipage}[t]{0.45\textwidth}
  \centering
    \includegraphics[width=0.22\textwidth]{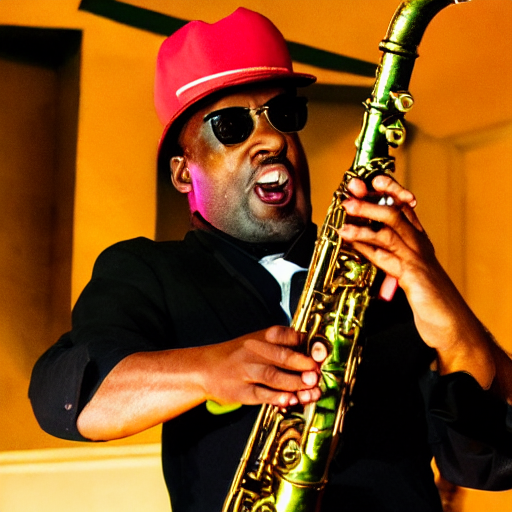}
    \includegraphics[width=0.22\textwidth]{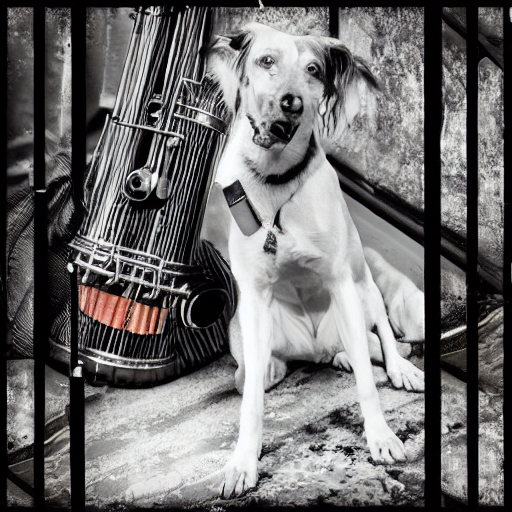}
    \includegraphics[width=0.22\textwidth]{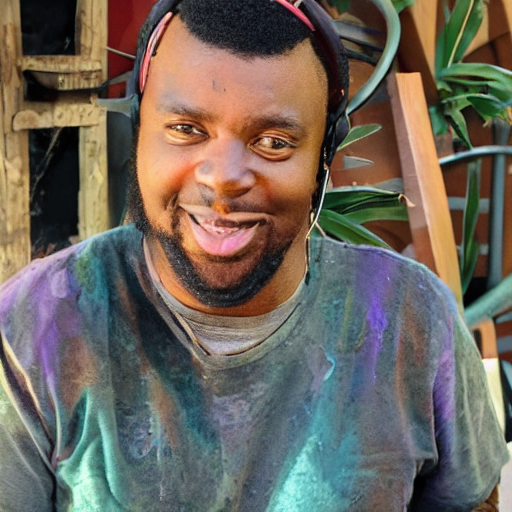}
    \includegraphics[width=0.22\textwidth]{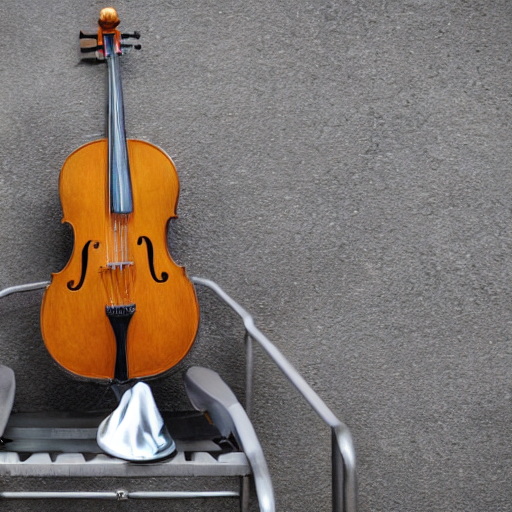}
    {\small Unlearning images regarding the music meaning of ``Jazz"}
  \end{minipage}
\hfill
\begin{minipage}[t]{0.45\textwidth}
  \centering
    \includegraphics[width=0.22\textwidth]{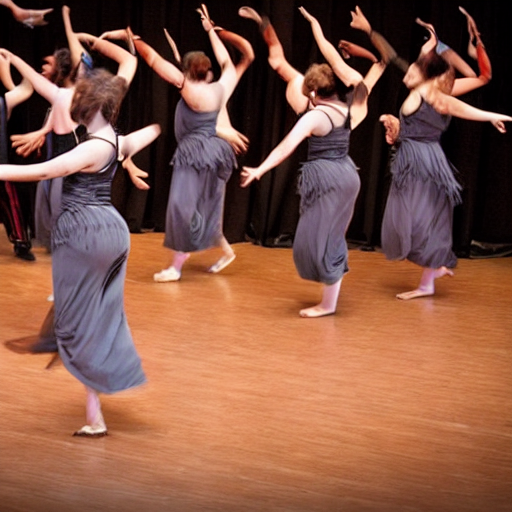}
    \includegraphics[width=0.22\textwidth]{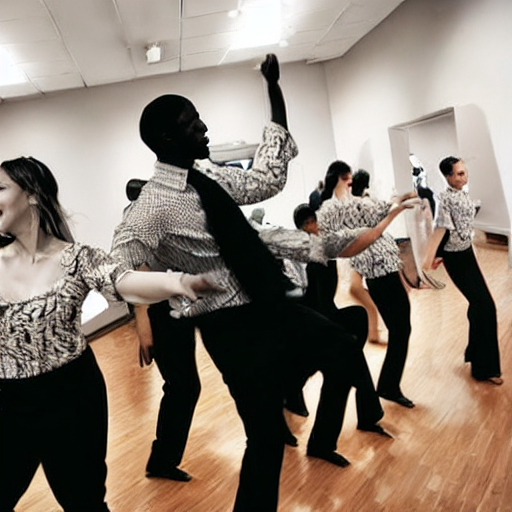}
    \includegraphics[width=0.22\textwidth]{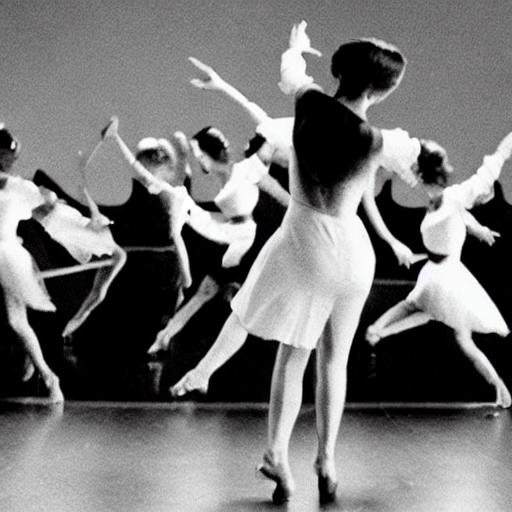}
    \includegraphics[width=0.22\textwidth]{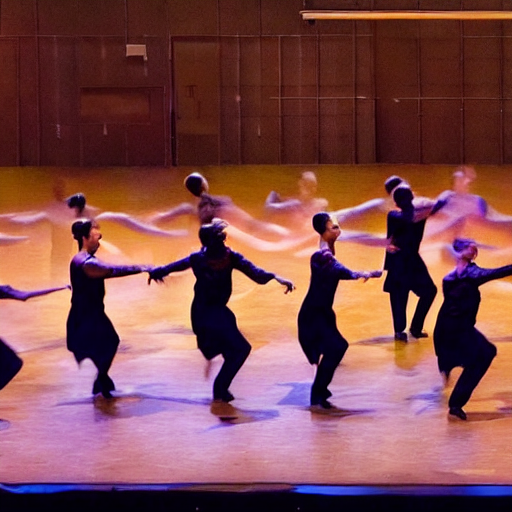}
    {\small Unlearning images regarding the dance meaning of ``Jazz"}
  \end{minipage}
  \hfill
    \begin{minipage}[t]{0.45\textwidth}
  \centering
    \includegraphics[width=\textwidth]{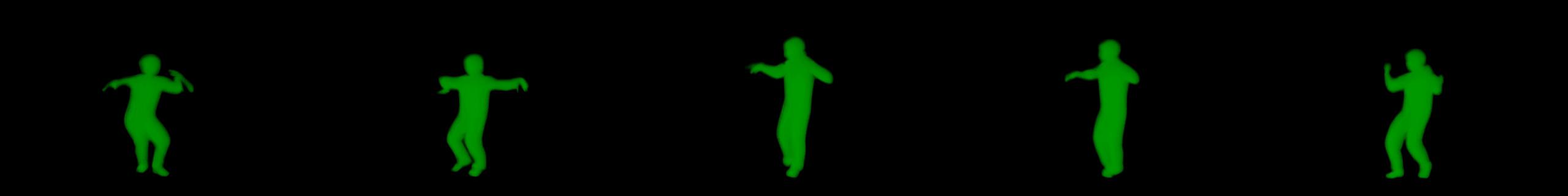}
    {\small Unlearn the music meaning of ``Jazz".}
  \end{minipage}
    \hfill
    \begin{minipage}[t]{0.45\textwidth}
  \centering
    \includegraphics[width=\textwidth]{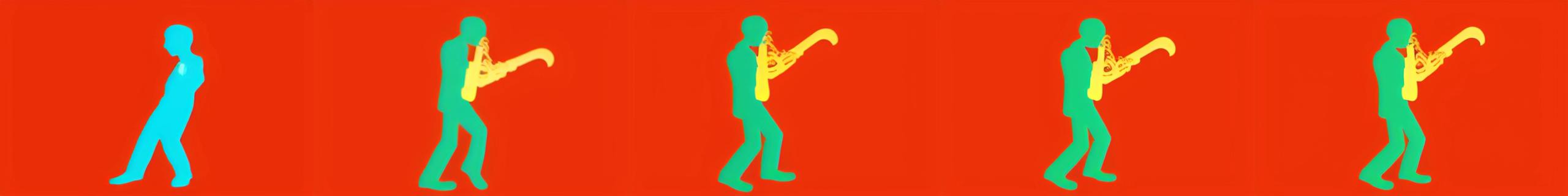}
    {\small Unlearn the dance meaning of ``Jazz".}
  \end{minipage}
  \caption{The comparison of unlearning the different meaning of polysemous concepts. \label{fig:multiple mearning}}
\end{figure*}

\subsection{Preliminary of Diffusion methods}
Diffusion models\cite{ho2020denoising,song2020denoising} are proposed to learn the underlying distribution of the data through diffusion and denoising process. For input data $\xb\sim p(\xb)$, in order to construct noisy sample $\xb_t=\alpha_t \xb + \sigma_t \varepsilonb$, the diffusion process adds random noise $\varepsilonb\sim \N(0,\Ib)$ to it to form a Markov Chain of T steps. The noise schedular is parameterized by the diffusion time step $t$, $\alpha_t$ and $\sigma_t$. Specifically, $SNR=\log(\alpha_t^2/\sigma_t^2)$, the signal-to-noise ratio monotonically decreases over time.  The models gradually denoise a normal distributed variable to learn the reverse process of the fixed Markov Chain of T steps to optimize a variational lower bound on $p(\xb)$. These models are weighted sequence of denoising U-Nets $\varepsilonb_{\thetab}(\xb,t)$ which predict the noises of $\xb_t$. The learning objective is as follows:
\begin{equation}\label{eqn:diffusion model}
  L_{DM}=\E_{\xb,\varepsilonb\sim \N(0,\Ib),t}\Vert \varepsilonb-\varepsilonb_{\thetab}(\xb_{t},t)\Vert.
\end{equation}

Latent diffusion models\cite{rombach2022high} use a variational autoencoder structure. The encoder $\Em$ compresses the input data into low-dimensional latent representation $\zb = \Em(\xb)$. Unlike direct diffusion models, the diffusion and denoising processes of latent diffusion models are implemented in the latent space. This setting saves substantial training and inference time. In the final denoising stage, the output is decoded as $\D(\zb_{0})$ which is the reconstructed data. The objective of latent diffusion models is as follows:
\begin{equation}\label{eqn:diffusion model}
  L_{LDM}=\E_{\Em(\xb),\varepsilonb\sim \N(0,\Ib),t}\Vert \varepsilonb-\varepsilonb_{\thetab}(\zb_{t},t)\Vert.
\end{equation}

The loss of latent diffusion models conditioned on the text input $\yb$ is as follows:
\begin{equation}\label{eqn:diffusion model}
  L_{LDM}=\E_{\Em(\xb),\yb,\varepsilonb\sim \N(0,\Ib),t}\Vert \varepsilonb-\varepsilonb_{\thetab}(\zb_{t},t,\cb_{\betab}(\yb))\Vert\
\end{equation}
where $\cb_{\betab}$ is the text encoder output and $\betab$ is the parameters of the text encoder.

\subsection{Unlearning method}
Because we are going to unlearn the concept in the text-to-image domain and transfer that unlearning capability to the text-to-video domain. The symbol $\xb$ discussed below represents images rather than videos, although videos are also feasible with significantly greater resource consumption in the following optimization.

In order to guarantee the semantic meaning of most of the concepts while changing only a small number of concepts, we apply a slight change to the text encoder's $\cb_{\betab}$ parameters
\begin{equation}\label{}
  \cb_{\betab}\leftarrow \cb_{\betab +\delta \betab}.
\end{equation}
 
In order to unlearn a specific concept $\yb^*$, a common unlearning method is to implement gradient ascent with respect to the parameters that can be optimized. The $\xb^*$ represent the images described the concept $\yb^*$. According to \cite{fuchi2024erasing},  we optimize the $\betab$, the parameters of the text encoder, with respect to the loss 
\begin{equation}\label{eqn:diffusion model}
  L_{LDM}=\E_{\Em(\xb^*),\yb,\varepsilonb\sim \N(0,\Ib),t}\Vert \varepsilonb-\varepsilonb_{\thetab}(\zb_{t},t,\cb_{\betab}(\yb^*))\Vert
\end{equation}
while keeping the other parameters frozen.

In order to make only slight changes to variables and minimize the impact on other concepts, we implement gradient ascent for only 5 epochs with respect to  $\xb^*$. $\xb^*$ is practically represented by 4 images to facilitate few-shot learning. These images are generated by text-to-image diffusion models to reduce the workload of collection.

After unlearning, we use the same text encoder, but we apply it for text-to-video generation. The procedure of our method is shown in Figure~\ref{fig:overview}.

\section{Experiment}
We conduct qualitative experiments and ablation studies.  Since there currently exists no other method for unlearning concepts in text-to-video diffusion models, we do not present a comparison of our method with other methods.
\subsection{Experiment Setup}
We apply our model on Lavie\cite{wang2023lavie} by transferring the text encoder, which is unlearned on the stable diffusion 1.5\cite{rombach2022high}. We use four generated images in the few shot setting to unlearn a specific concept. We optimize the text encoder using Adam\cite{kingma2014adam}. The hyperparameters used in the experiment are listed in Table~\ref{tab:hyperparameter}.
\begin{table}[h!]
\centering
 \caption{Hyperparamters\label{tab:hyperparameter}}
 \begin{tabular}{c c c c} 
 \hline
Hyperparameter & Value  \\  
 \hline
Training epochs & 5  \\ 
 Batch size & 2  \\
 Learning rate & $10^{-5}$  \\
 Weight decay & $10^{-8}$\\
 Adam$(\beta_1,\beta_2)$ & (0.9,0.98)  \\ 
 \hline
 \end{tabular}

\end{table}
\subsection{Qualitative Results}
We analyzed the qualitative results of unlearning a single object, its effect on other concepts, unlearning multiple concepts  and unlearning concepts with multiple meanings.

\subsubsection{Unlearning Single Object}
Single-object unlearning experiments are illustrated in Figure~1. 
We conducted the experiments on the concepts including ``Snoopy", ``Van Gogh", ``Eiffel Tower" and ``Elon Musk". Our method successfully replaced the Snoopy with a shadow of a wolf walking in the park at sunrise. Furthermore, our method removed the Van Gogh style orange background of a panda taking a selfie  and replaced it with a realistic green background. Additionally, for videos with the prompt ``a butterfly flying on the Eiffel Tower", our method removed the Eiffel Tower throughout the videos, leaving only a single butterfly flying in the frames. Finally, for the prompt ``Elon Musk in a space suit standing besides a rocket, high quality", our method removed the Elon Musk's facial features, protecting his privacy. These experiments demonstrated that our method can unlearn copyrighted cartoon characters, artist styles, objects and people's facial characteristics.

\subsubsection{Effect on Other Concepts}
We conducted an experiment to unlearn the concept ``Eiffel Tower". We compared the influence on the generation results of three other prompts. The results were shown in the Figure~\ref{fig:other effects}. There was no significant effect on the generations of ``a polar bear playing drum kit in NYC Times Square, 4k, high resolution" and ``jungle river at sunset, ultra quality". However, there seems to be an effect on the generation of the concept ``Tokyo Tower" in the prompt ``a cat climbing on Tokyo Tower". In the original generation, there was a Tokyo Tower in the scene. In the generation after unlearning ``Eiffer Tower", the Tokyo Tower seems to be interpreted as a general building. This experiment show that the unlearning procedure only influences the video generation of the concepts similar to target concepts and guarantees the generation quality of other concepts.

\subsubsection{Unlearning Multiple Concepts}
As people may be interested in unlearning multiple copyright concepts within a single text encoder, we conducted experiments focused on unlearning ``Snoopy" and ``Mickey Mouse." In cases without unlearning, the successful generations showed ``a Snoopy walking in the park at sunrise, oil painting style" and ``Mickey Mouse is greeting the children." When unlearning ``Snoopy," the first prompt's generation replaced Snoopy with the shadow of a wolf, while the second prompt successfully generated an image without Mickey Mouse's influence. When unlearning both ``Snoopy" and ``Mickey Mouse," the first prompt's generation again omitted Snoopy, and the second prompt's generation excluded Mickey Mouse. These experiments demonstrate the feasibility of unlearning multiple copyrighted concepts simultaneously.

\subsubsection{Unlearning Concepts with Multiple Meanings}
Since some concepts have multiple meanings and people want to unlearn a specific meaning of those concepts, we conducted experiments on unlearning ``Jazz". ``Jazz" has two meanings. One correlates with the music and is represented by the saxophone.  The other meaning is a dancing style. We aimed to unlearn the music meaning of the ``Jazz" by using several images related to saxophones and music performance. Similarly, we unlearn the dance meaning of ``Jazz" by using several images related to dance performance. To test the unlearning effect in the text-to-video domain, we used the prompt ``A person is dancing jazz." . The generation results of the unlearning process are shown in the Figure~\ref{fig:multiple mearning}. When unlearning the music meaning of ``Jazz", the video still showed the person dancing jazz. Conversely, when unlearning the dance meaning of ``Jazz", the videos showed  the person playing  saxophones, which is the characteristic of the music meaning of ``Jazz". These experiments demonstrate the possibility of unlearning specific meanings of the polysemous concepts which are copyright-related.

Because there is no proper metric for evaluating unlearning performance, we did not implement quantitative experiments.
\subsection{Ablation Studies}
Our method involves the $k$-shot unlearning process in the text-to-image text encoder and the training epoch number of the unlearning process. In the following experiment, we demonstrate the influence of different settings for transfer learning  on the text-to-video generation. 
\subsubsection{$k$-shot Unlearning}
When implementing the unlearning, we need to collect several concept-related images to help unlearn the concept of the text encoder and then to transfer the unlearning effect to the text-to-video diffusion models. Normally, we use the four generated images as the concept-related images and it is the four-shot learning. In this experiment, we compare zero-shot, two-shot and four-shot cases.

\begin{figure}[h]
\centering

\begin{subfigure}{0.45\textwidth}
        \centering
        \includegraphics[width=\linewidth]{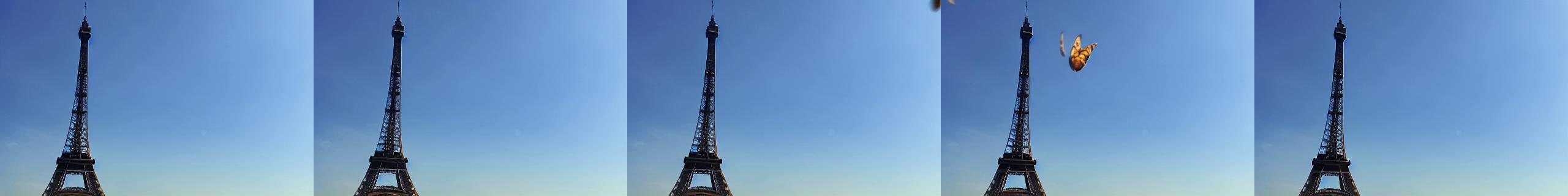}
        \caption{Unlearn "Eiffel Tower" zero-shot}
        \label{fig:subfig1}
    \end{subfigure}
\begin{subfigure}{0.45\textwidth}
        \centering
        \includegraphics[width=\linewidth]{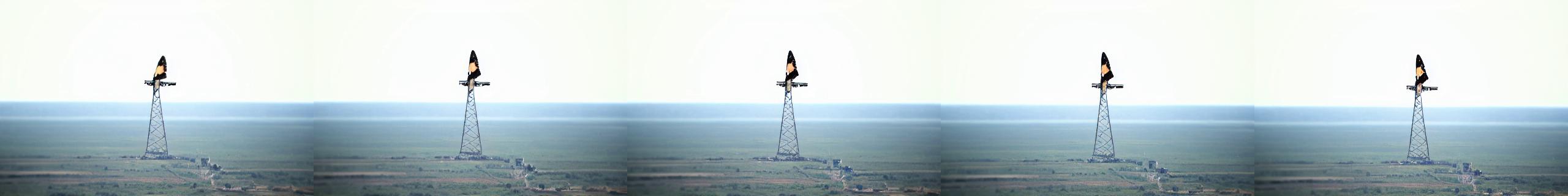}
        \caption{ Unlearn "Eiffel Tower" two-shot}
        \label{fig:subfig2}
    \end{subfigure}
\begin{subfigure}{0.45\textwidth}
        \centering
        \includegraphics[width=\linewidth]{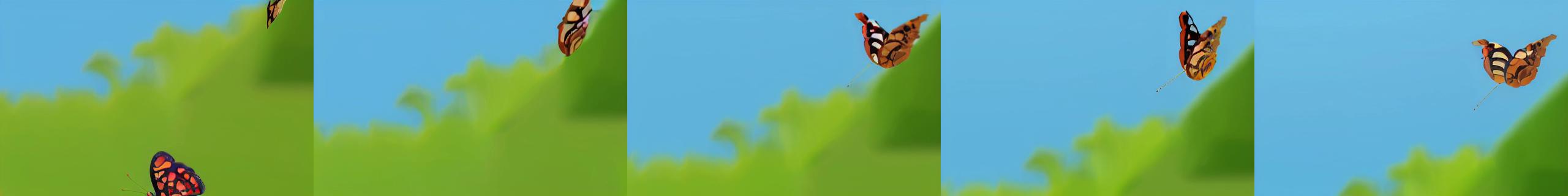}
        \caption{  Unlearn "Eiffel Tower" four-shot}
        \label{fig:subfig3}
    \end{subfigure}
\caption{The comparison of the effects  after unlearning ``Eiffel Tower" in different shots. The prompt is ``a butterfly flying on the Eiffel Tower". \label{fig:k-shot} }
\end{figure}
As shown in the Figure~\ref{fig:k-shot}, the concepts and images of Eiffel Tower were gradually removed as the number of shots increased. During unlearning ``Eiffel Tower" with zero-shot images, the Eiffel Tower concepts and images remained preserved in the video. When unlearning ``Eiffel Tower" with two-shot images, the video depicted the Eiffel Tower  as a high-voltage power tower. It seems that the high-voltage power tower is similar to the Eiffel Tower. Finally, with four-shot images, the Eiffel Tower was completely removed from the video.

\subsubsection{Number of Epochs}
To implement unlearning, we need to train for several epochs to unlearn the concepts learned by the text encoder. Then, we transfer this unlearning effect to the text-to-video diffusion models. Typically, five epochs are used for training. In this experiment, we compare the training results obtained with one, two, three, four, and five epochs.

\begin{figure}[h]
\centering

\begin{subfigure}{0.45\textwidth}
        \centering
        \includegraphics[width=\linewidth]{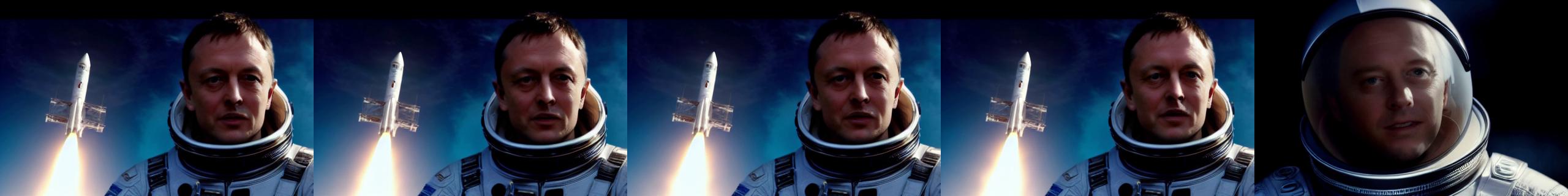}
        \caption{Epoch 1}
        \label{fig:Epoch1}
    \end{subfigure}
\begin{subfigure}{0.45\textwidth}
        \centering
        \includegraphics[width=\linewidth]{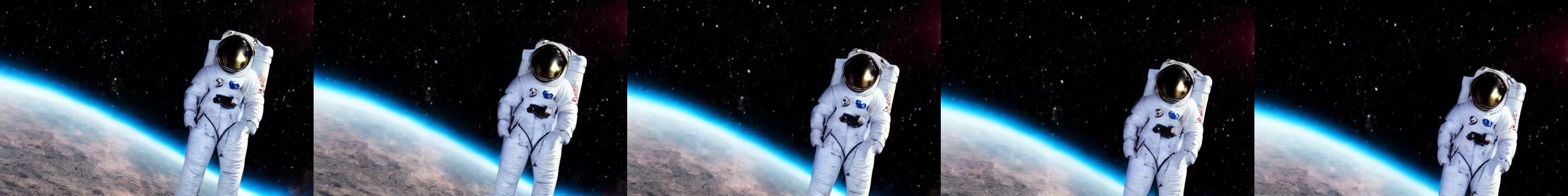}
        \caption{ Epoch 2}
        \label{fig:Epoch2}
    \end{subfigure}
\begin{subfigure}{0.45\textwidth}
        \centering
        \includegraphics[width=\linewidth]{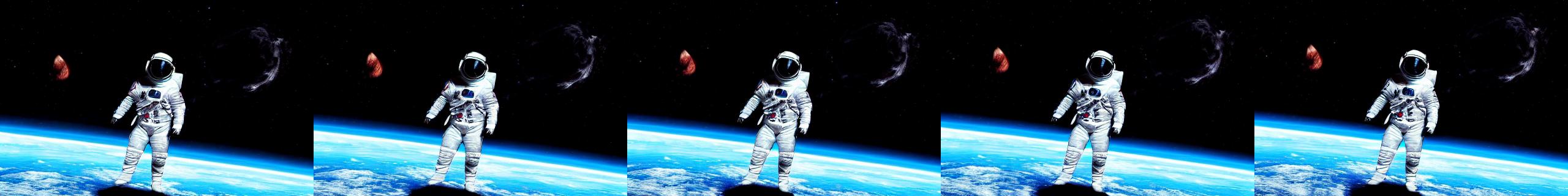}
        \caption{ Epoch 3}
        \label{fig:Epoch3}
    \end{subfigure}
\begin{subfigure}{0.45\textwidth}
        \centering
        \includegraphics[width=\linewidth]{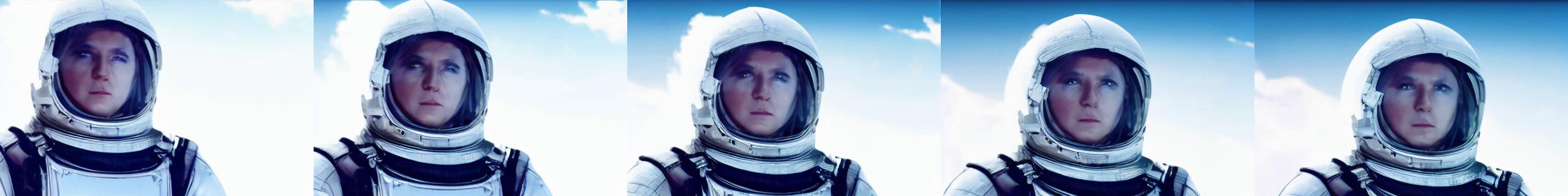}
        \caption{ Epoch 4}
        \label{fig:Epoch4}
\end{subfigure}
\begin{subfigure}{0.45\textwidth}
        \centering
        \includegraphics[width=\linewidth]{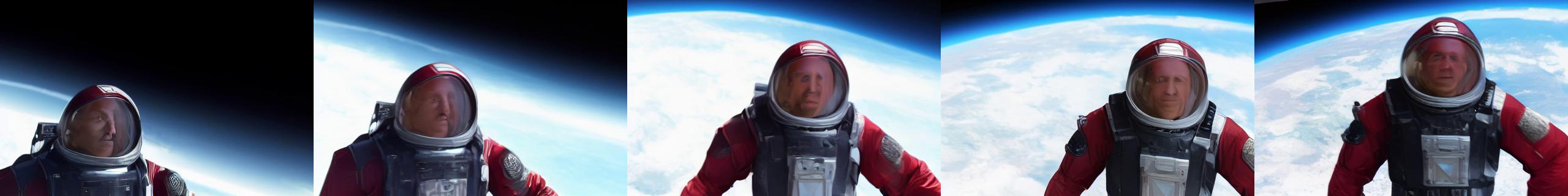}
        \caption{ Epoch 5}
        \label{fig:Epoch5}
\end{subfigure}
\caption{The comparison of the effects  after unlearning ``Elon Musk" in different epoch trainings. The prompt is ``Elon Musk in a space suit standing besides a rocket, high quality". \label{fig:epoch_number}}
\end{figure}
As we can see in Figure~\ref{fig:epoch_number}, with an increasing number of epochs, Elon Musk's facial characteristics disappear.
When the epoch number is 1, the generated man's facial characteristics are similar to Elon Musk. After epoch number 2, Elon Musk's facial characteristics disappear, and we cannot tell the identity of the generated man.

\section{Limitations}
The experiments demonstrate the effectiveness of our transfer method. However, based on previous experiments, we observed that when the model unlearns a specific concept, the generation of similar concepts can be influenced. For instance, when unlearning "Eiffel Tower," the generation of "Tokyo Tower" was affected, and the semantic meaning of "Tokyo Tower" seemed to be mapped to "building." Furthermore, we found that when a specific concept is unlearned, the new meaning associated with the concept name may not be significantly similar to its original semantic meaning. This contradicts the conclusion presented in \cite{fuchi2024erasing}.

\section{Conclusion}
Our method achieves unlearning concepts  by transferring the unlearning capability of the text encoder from text-to-image diffusion models to text-to-video diffusion models. The process of unlearning a specific concept involves optimizing the parameters of the text encoder through gradient ascent on the objective function based on several concept-related images. The optimized text encoder is then reused  in the text-to-video diffusion models. Since the unlearning procedure focuses on the text-to-image domain and only optimizes the text encoder, it is fast, taking only about 100s to unlearn a concept on an RTX 3070. In our experiment, we found that several target concepts disappeared in the videos while the effect on other, different concepts was minimal. The models can also jointly unlearn multiple concepts. Additionally, our ablation study demonstrate the need for a sufficient number of images for few-shot learning and for the models to be optimized by a sufficient number of epochs.

Our method can unlearn static concepts like objects, artistic styles, cartoon characters, and human appearances. However, it cannot handle dynamic concepts such as specific dance routines, acrobatics, and continuous photographic works. These are difficult to express in the text-to-image domain and are also protected by copyright law. We plan to investigate how to unlearn dynamic concepts in the future. This may involve directly optimizing the text encoder for the text-to-video domain. Ultimately, we aim to combine optimizations in both text-to-image and text-to-video domains to achieve unlearning of dynamic concepts.

\bibliography{aaai25}

\end{document}